\documentclass{article}
\usepackage{ijcai25}

\usepackage{times}
\usepackage{soul}
\usepackage[hidelinks]{hyperref}
\usepackage[utf8]{inputenc}
\usepackage[small]{caption}
\usepackage{graphicx}
\usepackage{amsmath,amssymb,amsfonts}
\usepackage{amsthm}
\usepackage{booktabs}
\usepackage{algorithm}
\usepackage{algorithmic}
\usepackage[switch]{lineno}
\usepackage{color}
\usepackage{amssymb}
\usepackage{subcaption}
\usepackage{caption}
\usepackage{multirow}

\title{FancyVideo: Towards Dynamic and Consistent Video Generation via Cross-frame
Textual Guidance}

\author{
Jiasong Feng$^{1,2,*}$\and
Ao Ma$^{1,3,*,\dag}$\and
Jing Wang$^{1,4,*}$\and
Ke Cao$^{1,5,*}$\And
Zhanjie Zhang$^{1,6,\ddag}$
\\
\affiliations
$^1$ 360 AI Research \\
$^2$ Beijing University of Technology\\
$^3$ Wuhan University\\
$^4$ Sun Yat-sen University\\
$^5$ University of Science and Technology of China\\
$^6$ Zhejiang University\\
\emails
\{maaoaoma, zhangzhanj\}@126.com
}

\begin{document}
\maketitle
\let\thefootnote\relax

\footnotetext{$^*$ Equal contribution.}
\footnotetext{$^\dag$ Project leader.}
\footnotetext{$^\ddag$ Corresponding author.}

\begin{abstract}

Synthesizing motion-rich and temporally consistent videos remains a challenge in artificial intelligence, especially when dealing with extended durations. Existing text-to-video (T2V) models commonly employ spatial cross-attention for text control, equivalently guiding different frame generations without frame-specific textual guidance.  
Thus, the model's capacity to comprehend the temporal logic conveyed in prompts and generate videos with coherent motion is restricted. 
To tackle this limitation, we introduce \textbf{FancyVideo}, an innovative video generator that improves the existing text-control mechanism with the well-designed \textbf{C}ross-frame \textbf{T}extual \textbf{G}uidance \textbf{M}odule (CTGM). 
Specifically, CTGM incorporates the Temporal Information Injector (TII) and Temporal Affinity Refiner (TAR) at the beginning and end of cross-attention, respectively, to achieve frame-specific textual guidance. Firstly, TII injects frame-specific information from latent features into text conditions, thereby obtaining cross-frame textual conditions. Then, TAR refines the correlation matrix between cross-frame textual conditions and latent features along the time dimension.  
Extensive experiments comprising both quantitative and qualitative evaluations demonstrate the effectiveness of FancyVideo. Our approach achieves state-of-the-art T2V generation results on the EvalCrafter benchmark and facilitates the synthesis of dynamic and consistent videos. Note that the T2V process of FancyVideo essentially involves a text-to-image step followed by T+I2V. This means it also supports the generation of videos from user images, i.e., the image-to-video (I2V) task. A significant number of experiments have shown that its performance is also outstanding. 
\end{abstract}

\section{Introduction}
With the advancement of the diffusion model, the text-to-image (T2I) generative models \cite{blattmann2023align,ho2022imagen,luo2023decomposed,ma2024hico,liu2025bridge,cao2025relactrl,ling2025ragar,bi2024using,ma2025nami} can produce high-resolution and photo-realistic images by complex text prompts, resulting in various applications. Currently, many studies \cite{wang2024qihoo,guo2023i2v} explore the text-to-video (T2V) generative model due to the great success of T2I models. However, building a powerful T2V model remains challenging as it requires maintaining temporal consistency while generating coherent motions simultaneously. 
Moreover, due to limited memory, most diffusion-based T2V models \cite{wang2024qihoo,guo2023i2v,zhang2024moonshot,guo2023animatediff,chen2023seine,menapace2024snap} can only produce fewer than 16 frames of video per sampling without extra assistance (i.e., super-resolution).

\begin{figure*}[ht]
  \centering
  \includegraphics[width=1\linewidth]{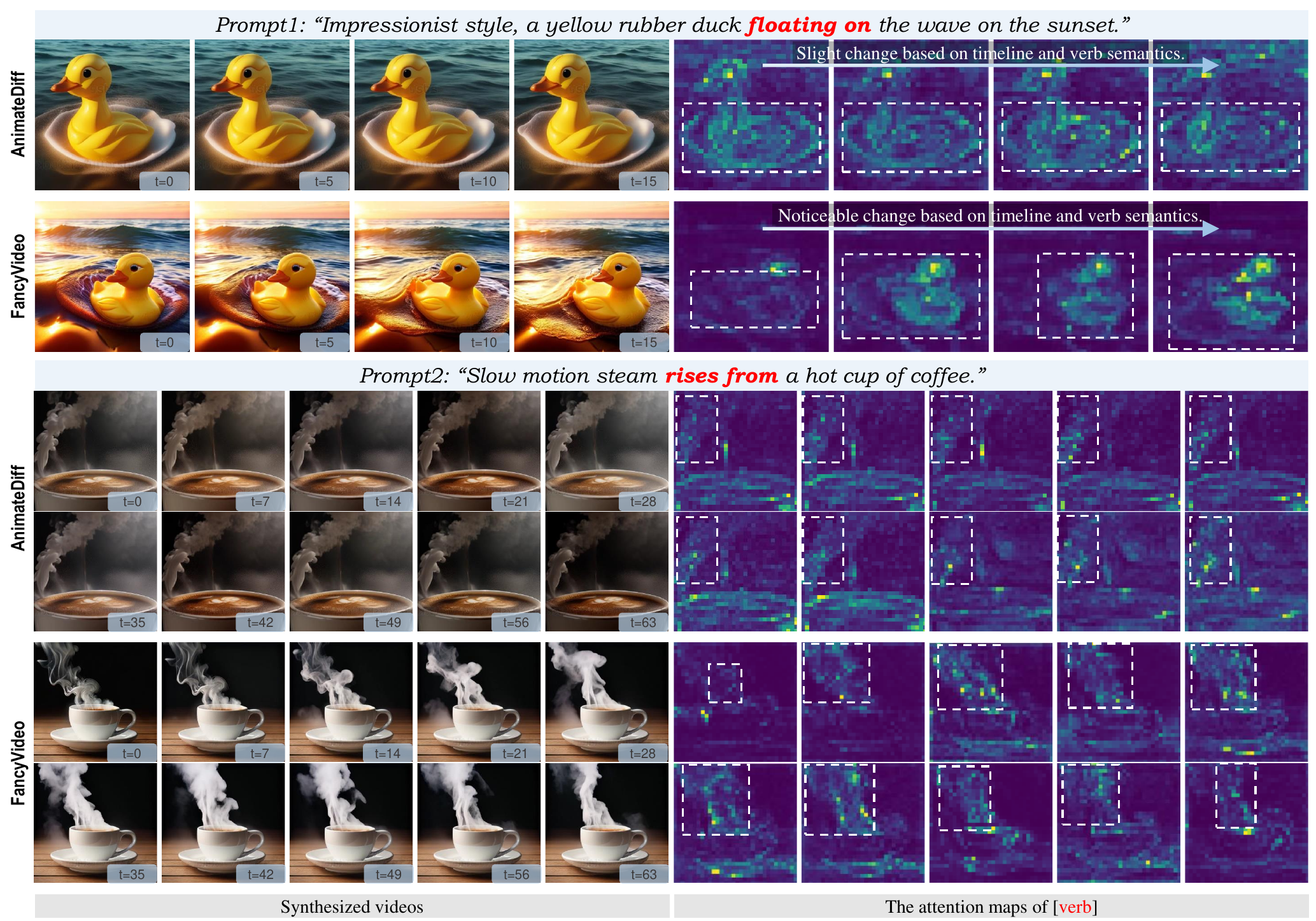}
  \caption{The generated videos and the attention maps of [\textcolor{red}{verb}] belong to FancyVideo and AnimateDiff. We present the 16-frame video (top) and longer 64-frame video (bottom). Due to the inadequate time-specific textual guidance in the AnimateDiff, the [\textcolor{red}{verb}] focused region remains almost constant, resulting in a lack of motion in the video. 
In contrast, \textbf{FancyVideo} effectively alleviates this issue through cross-frame textual guidance. The [\textcolor{red}{verb}] focused region changes based on the timeline and semantics, thereby generating motion-rich videos.}

  \label{fig1}
\end{figure*}

\begin{figure}[htbp]
\centering
\includegraphics[width=0.45\textwidth]{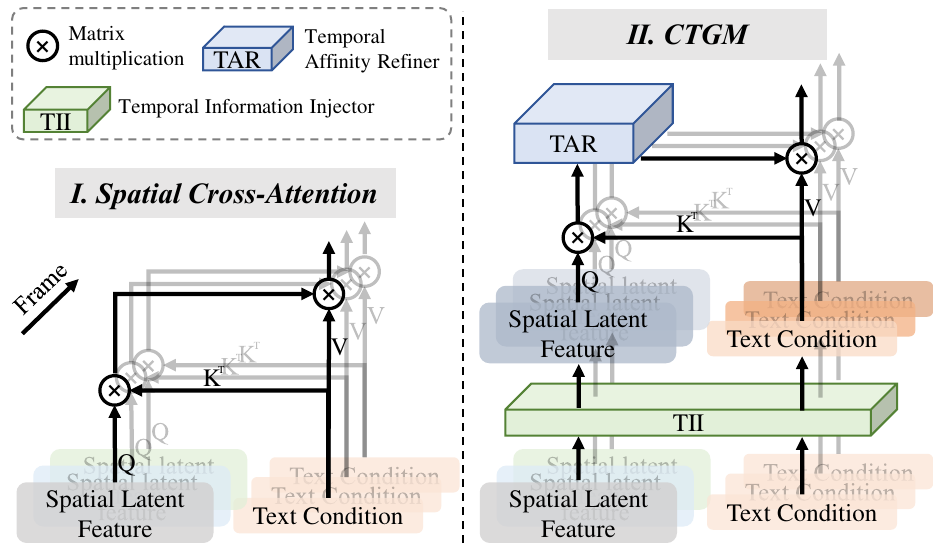}
\caption{\small
The structure of 
spatial cross-attention and CTGM.}
\label{CTGM}
\end{figure}

The existing T2V models \cite{zhang2024moonshot,guo2023animatediff,chen2023seine,menapace2024snap,bi2025customttt,wang2025wisa,shao2025eventvad} typically employ spatial cross-attention between text conditions and latent features for achieving text control generation. However, as shown in Fig. \ref{CTGM}(\uppercase\expandafter{\romannumeral1}), this manner shares the same text condition across different frames, thus lacking the specific textual guidance tailored to each frame. 
Consequently, these T2V models struggle to comprehend the temporal logic of text prompts and produce videos with coherent motion. 
Taking AnimateDiff \cite{guo2023animatediff} as an example, in  Fig. \ref{fig1}, we exhibit its generated video and visualize the [verb]-focused region (which is closely associated with the video motion) based on the attention map from the cross-attention module.
Ideally, these regions should transition smoothly over time and align with the semantics of motion instructions. However, as observed in the upper right of the figure, the [verb]-focused region remains nearly identical across different frames due to the consistent textual guidance between frames. Meanwhile, the video exhibits poor motion in the upper left of the figure.

Furthermore, we perform a similar visual analysis for the longer video (e.g., 64 frames) generation and find that this problem is more prominent, as illustrated in the lower part of Fig. \ref{fig1}. 
Therefore, we believe this approach hampers the advancement of video dynamics and consistency and is sub-optimal for video generation tasks based on text prompts.

To this end, we present a novel T2V model named \textbf{FancyVideo}, capable of comprehending complex spatial-temporal relationships within text prompts. By employing a cross-frame textual guidance strategy, FancyVideo can generate more dynamic and plausible videos in a sampling process. 
Specifically, to boost the model's capacity for understanding spatial-temporal information in text prompts, we optimize the spatial cross-attention through the proposed \textbf{C}ross-frame \textbf{T}extual \textbf{G}uidance \textbf{M}odule (CTGM), comprising a Temporal Information Injector (TII) and Temporal Affinity Refiner (TAR). 
As illustrated in Fig. \ref{CTGM}(\uppercase\expandafter{\romannumeral2}), TII injects temporal information from latent features into text conditions, building cross-frame textual conditions. Then, TAR refines the affinity between frame-specific text embedding and video along time dimension, adjusting the temporal logic of textual guidance. 
Through the cooperative interaction between TII and TAR, FancyVideo fully captures the motion logic embedded within images and text. Consequently, its motion token-focused area shifts logically with frames, as illustrated in the lower right part of Fig. \ref{fig1}. 
This characteristic enables FancyVideo to produce dynamic videos, as displayed in the lower left part of the figure. 
Experiments demonstrate that FancyVideo successfully generates dynamic and consistent videos, achieving the SOTA results on the EvalCrafter \cite{liu2023evalcrafter} benchmark and the competitive performance on UCF-101 \cite{soomro2012dataset} and MSR-VTT \cite{xu2016msr}. Additionally, FancyVideo supports generating videos from user-input images, i.e., the image-to-video task. We have also conducted extensive experiments to demonstrate the superiority of our method.

\paragraph{Contributions.}
\textbf{1)} We introduce FancyVideo, the pioneering endeavor as far as our knowledge extends, delving into cross-frame textual guidance for the T2V task. 
This approach offers a fresh perspective to enhance current text-control methodologies.
\textbf{2)} We propose the Cross-frame Textual Guidance Module (CTGM), which constructs cross-frame textual conditions and subsequently guides the modeling of latent features with robust temporal plausibility.
It can effectively enhance the motion and consistency of video.
\textbf{3)} We demonstrate that incorporating cross-frame textual guidance represents an effective approach for achieving high-quality video generation.
Our experiments showcase that this approach attains state-of-the-art results on both quantitative and qualitative evaluations.

\section{Related Work}
    \paragraph{Text to Video Generation.} Generative models like GANs \cite{wang2020imaginator,munoz2021temporal,gur2020hierarchical}, auto-regressive models \cite{wang2019point,yan2021videogpt}, and implicit neural representations \cite{de2023deep} have been explored for video generation. Recently, diffusion models \cite{rombach2022high,zhang2024towards,zhang2024artbank} have advanced text-to-image quality. Stable Diffusion \cite{rombach2022high} uses a VAE \cite{kingma2013auto} latent space to reduce cost \cite{jiang2023res}. T2V models \cite{wu2023tune} add temporal layers to T2I models but often lack frame-to-frame consistency. We propose cross-frame textual guidance to improve temporal coherence.

\paragraph{Image-conditioned Video Generation.}
To bridge the gap between text and video, recent work leverages images for clearer video generation. SVD \cite{blattmann2023stable} treats images as noisy latent inputs, while MoonShot \cite{zhang2024moonshot} improves semantic consistency using a CLIP encoder. Though effective, these I2V methods rely on input images. Hierarchical approaches \cite{zeng2023make,chen2023seine} use images as keyframes to extend video length with fewer constraints. These methods, though I2V-capable, are essentially T2V. FancyVideo adopts a hierarchical design with cross-frame textual guidance, enabling more frames per iteration and faster inference.

\section{Method}
\begin{figure*}[ht]
  \centering
  \includegraphics[width=1\linewidth]{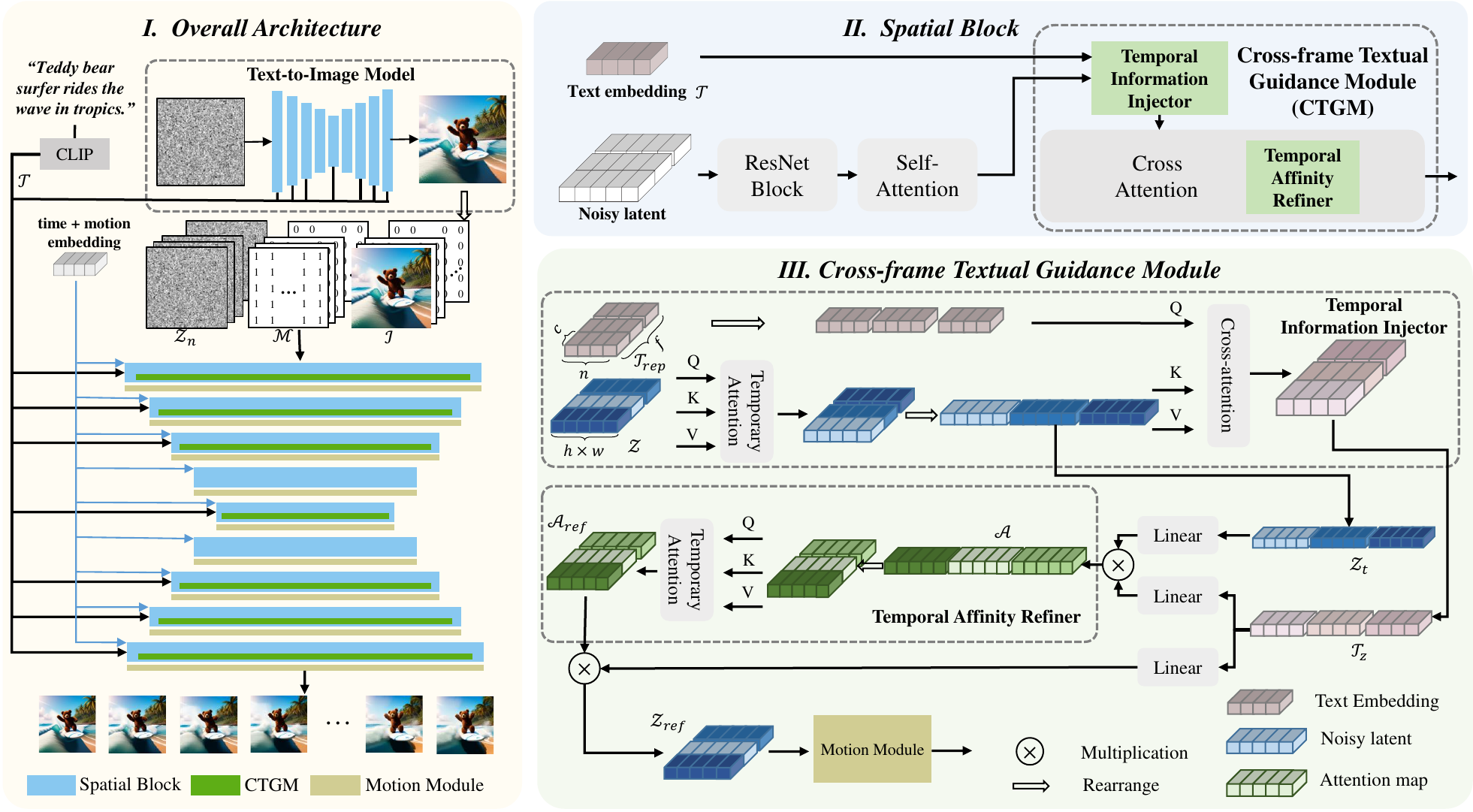}
  \caption{The overall architecture of our method. FancyVideo is a T+I2V model that concatenates noise latent, mask indicator, and image indicator as input. We insert our Cross-frame Textual Guidance Module (CTGM) into each spatial block. CTGM consists of three components: Temporal Information Injector, Temporal Affinity Refiner, and Temporal Feature Booster (see supplementary materials). These components are inserted at the beginning, middle, and end of cross-attention, respectively.}

  \label{pipeline_new}
\end{figure*}
\label{sec:method}
\subsection{Preliminaries}

\paragraph{Latent Diffusion Models.}
LDMs \cite{sohl2015deep,ho2020denoising,zhang2025u,he2025plangen,lu2025uni} enhance efficiency by running diffusion in the VAE-compressed latent space \cite{kingma2013auto} instead of pixel space. The forward process adds Gaussian noise ($\epsilon \sim \mathcal{N}(0, I)$) to the latent code $\mathbf{z}$, yielding:
\begin{equation}
\mathbf{z}_t=\sqrt{\bar{\alpha_t}}\mathbf{z}+\sqrt{1-\bar{\alpha_t}} \epsilon,
\end{equation}
where $\bar{\alpha_t}$ denotes a noise scheduler with timestep $t$. For the inverse process, it trains a denoising model ($f_\theta$) with the objective:
\begin{equation}
\mathbb{E}_{\mathbf{z}\sim p(z), \epsilon\sim\mathcal{N}(0,I), t}\left[\left\|\boldsymbol{y}-f_\theta(\mathbf{z}_t,\mathbf{c},t)\right\|^2\right],
\end{equation}
where $\mathbf{c}$ represents the condition and target $\boldsymbol{y}$ can be noise $\epsilon$, denoising input $\mathbf{z}$ or $\boldsymbol{v}$-prediction ($\boldsymbol{v}= \sqrt{\bar{\alpha_t}} \epsilon - \sqrt{1-\bar{\alpha_t}}\mathbf{z}$) in \cite{salimans2022progressive}. In this paper, we adopt the $\boldsymbol{v}$-prediction as the supervision.

\paragraph{Zero terminal-SNR Noise Schedule.} 
Previous studies proposed zero terminal SNR \cite{lin2024common} to handle the signal-to-noise ratio (SNR) difference between the testing and training phase, which hinders the generation quality.
At training, due to the residual signal left by the noise scheduler, the SNR is still not zero at the terminal timestep $T$.
However, the sampler lacks realistic data when sampling from random gaussian noise during the test, resulting in a zero SNR.
This train-test discrepancy is unreasonable and an obstacle to generating high-quality videos. Therefore, following the \cite{lin2024common,girdhar2023emu}, we scale up the noise schedule and set $\bar{\alpha_T}=0$ to fix this problem.

\subsection{Model Architecture}

Fig. \ref{pipeline_new} illustrates the overall architecture of FancyVideo. The model is structured as a pseudo-3D UNet, which integrates frozen spatial blocks, sourced from a text-to-image model, along with Cross-frame Textual Guidance Modules (CTGM) and temporal attention blocks. The model takes three features as input: noisy latent $\mathcal{Z}_n \in \mathbb{R}^{ f \times h \times w \times c}$, where $h$ and $w$ indicate the height and width of the latent, $f$ signifies the number of frames, and $c$ denotes the channels of the latent; mask indicator $\mathcal{M} \in \mathbb{R}^{ f \times h \times w \times 1}$, with elements set to 1 for the first frame and 0 for all other frames; image indicator $\mathcal{I} \in \mathbb{R}^{ f \times h \times w \times c}$, with initial image as the first frame and 0 for all other frames. The denoising input $\mathcal{Z}$ is formed by concatenating $\mathcal{Z}_n$, $\mathcal{M}$ and $\mathcal{I}$ along the channel dimension, represented as $\mathcal{Z}=[\mathcal{Z}_n;\mathcal{M};\mathcal{I}] \in \mathbb{R}^{ f \times h \times w \times (2c+1)}$.  Within each spatial block, we first incorporate prior knowledge of the motion score as embeddings. In each subsequent cross-attention layer, CTGM is employed to capture the intricate dynamics described in the text prompts. Afterward, we apply temporal attention blocks to enhance the temporal relationships across various patches.

\subsubsection{Motion Embedding}
To achieve more controllable video generation in terms of motion amplitude, we introduce motion score information calculated by the RAFT \cite{teed2020raft} alongside the timestep information. Specifically, we calculate a motion score for the training samples in the dataset within a range of 0.1 to 10. The score are then encoded into motion features through a motion embedding layer. By controlling the motion score, we can generate videos with stronger motion. However, simply adjusting the score may lead to unrealistic motion. We use CTGM to prevent these issues.

\subsubsection{Cross-frame Textual Guidance Module}

CTGM advances the existing text control method through two sub-modules: Temporal Information Injector (TII) and Temporal Affinity Refiner (TAR) as depicted in Fig. \ref{pipeline_new}(\uppercase\expandafter{\romannumeral3}). Before engaging in cross-attention, TII initially extracts temporal latent feature $\mathcal{Z}_t$ and then incorporates temporal information into text embedding $\mathcal{T}_{rep}$ based on $\mathcal{Z}_t$, obtaining cross-frame textual condition $\mathcal{T}_z$.  Subsequently, TAR refines the affinity between $\mathcal{Z}_t$ and  $\mathcal{T}_z$ along the time axis, enhancing the temporal coherence of textual guidance.
The computation process of the CTGM can be formalized as:
\begin{align}
\mathcal{Z}_t, \mathcal{T}_z &=\mathrm{TII}(\mathcal{Z}, \mathcal{T}_{rep}), \\
\mathcal{Z}_{ref}=\mathrm{Softmax}(&\frac{\mathrm{TAR}(W_q\mathcal{Z}_t , W_k\mathcal{T}_z)}{\sqrt{d_k} }W_v(\mathcal{T}_z),
\end{align}
where $W_q$, $W_k$, and $W_v$ represent the linear layers for query, key, and value in original cross-attention, respectively. The hyper-parameter $d_k$ is acquired from the query dimensions. $\mathrm{TII}(\cdot, \cdot)$ and $\mathrm{TAR}(\cdot)$ denotes the functions of TII and TAR. In the end, we get refined noisy latent feature ${Z}_{ref}$. A detailed description of these three modules is provided as follows.

\paragraph{Temporal Information Injector.}
In previous work \cite{guo2023animatediff,girdhar2023emu}, the text embedding $\mathcal{T}_{rep}$ is repeated equally $f$ times, resulting in $\mathcal{T}_{rep} \in \mathbb{R}^{ f \times n \times c}$, $n$ denoting the length of the embedding vector. We inject temporal information into the embedding before performing spatial cross-attention, thereby enabling distinct focal points on the text within different frames. In Temporal Information Injector (TII), we initially reshape the noisy latent $\mathcal{Z}$ from $\mathbb{R}^{ f \times h \times w \times c}$ to $\mathbb{R}^{ (h w) \times f \times c}$ and apply temporal self-attention to acquire $\mathcal{Z}_t$. Then, we conduct spatial cross-attention, using the repeated text embedding $\mathcal{T}_{rep}$ as queries and the noisy latent $\mathcal{Z}_t \in \mathbb{R}^{ f \times (h w) \times c}$ as both keys and values, resulting in the text embedding $\mathcal{T}_z$ with frame-specific temporal information. The formalization of the TII module can be expressed as follows:
\begin{equation}
\begin{split}
\mathcal{Z}_t, \mathcal{T}_z &= \mathrm{TII}(\mathcal{Z}, \mathcal{T}_{rep}) \\
&= \mathrm{SelfAttn_t}(\mathcal{Z}), \\
&\quad \mathrm{CrossAttn_s}(\mathrm{SelfAttn_t}(\mathcal{Z}), \mathcal{T}_{rep})
\end{split}
\end{equation}
where $\mathrm{SelfAttn_t}$ denotes temporal self-attention and $\mathrm{CrossAttn_s}$ denotes spatial cross-attention. Through TII, we obtain the noisy latent $\mathcal{Z}_t$ with temporal information and the latent-aligned text embedding $\mathcal{T}_z$.

\paragraph{Temporal Affinity Refiner.}
To dynamically allocate attention to text embedding across different frames, we design the Temporal Affinity Refiner (TAR) to refine the attention map of spatial cross-attention. In spatial cross-attention, the noisy latent serves as the query, while the text embedding serves as both the key and value. The attention map $\mathcal{A} \in \mathbb{R}^{ f \times (h w) \times n}$, compute as $\mathcal{A}=(W_q \mathcal{Z}_t)  (W_k \mathcal{T}_z)^{T} / \sqrt{d_k}$, reflects the affinity between the text and patches. Then, TAR applies temporal self-attention to the attention map $\mathcal{A} \in \mathbb{R}^{ (h w) \times f \times n}$, obtaining the refined attention map $\mathcal{A}_{ref}$, which can be represented as:

\begin{equation}
\mathcal{A}_{ref} = \mathrm{TAR}(\mathcal{A}) = \mathrm{SelfAttn_t}(\mathcal{A})
\end{equation}
With the TAR, $\mathcal{A}_{\text{ref}}$ establishes a more logical temporal connection in the affinity matrix. It can perform more dynamic action while ensuring no additional video distortion occurs. Finally, the cross-attention process is completed with the refined attention map as $\mathcal{Z}_{ref}=\mathrm{Softmax}(\mathcal{A}_{ref})(W_v \mathcal{T}_z)$.

\begin{figure*}[!htbp]
  \centering
  \includegraphics[width=0.87\linewidth]{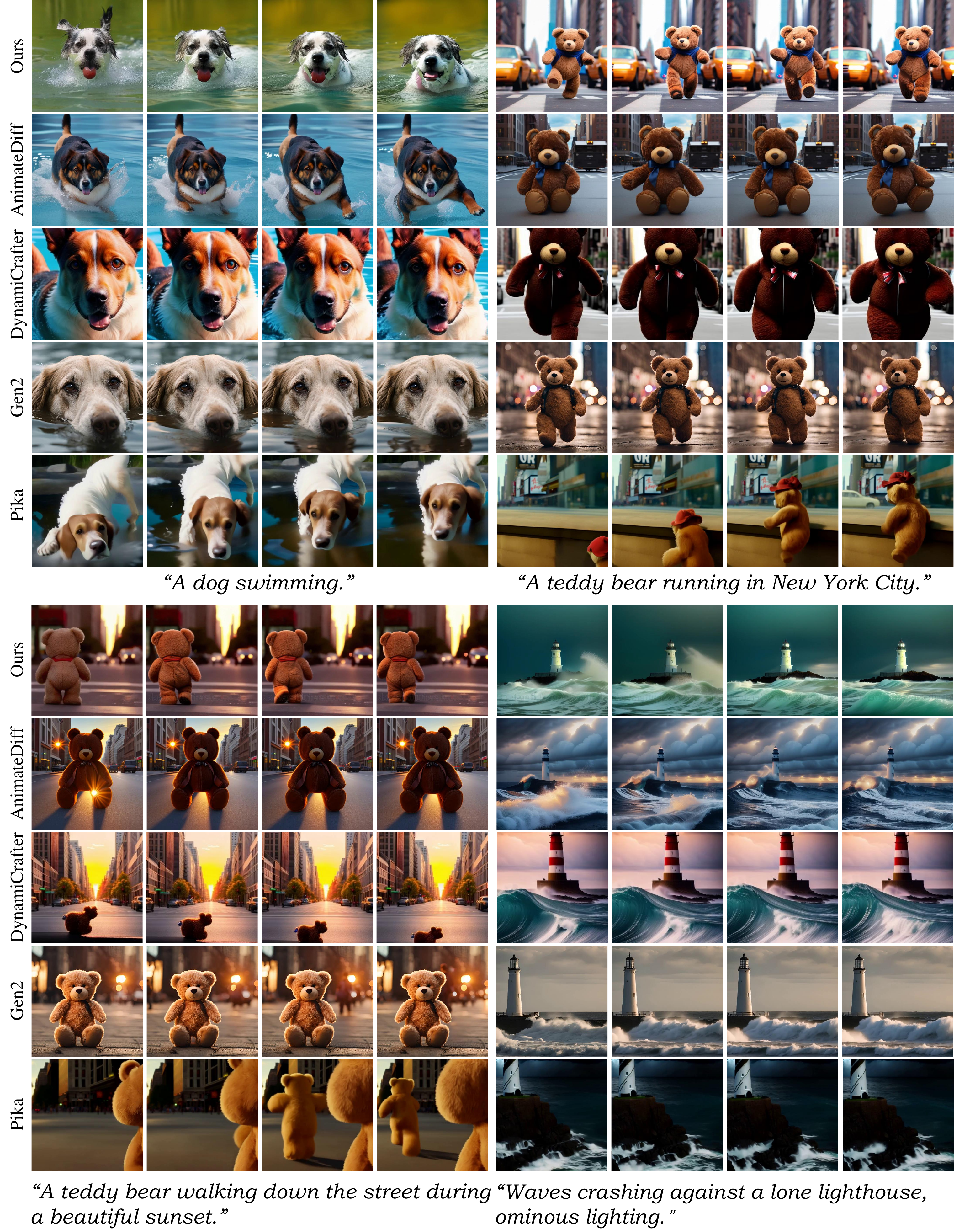}
  \caption{Qualititive analysis. 
  We compare the video generation results from AnimateDiff~\protect\cite{guo2023animatediff}, DynamiCrafter~\protect\cite{xing2023dynamicrafter}, Pika~\protect\cite{pikalab2023}, Gen-2~\protect\cite{gentwo2023}, and our FancyVideo.
  }
  \label{Qualitative_eval}
\end{figure*}

\section{Experiments}
\label{sec:conclusion}

In the quantitative experiments, FancyVideo utilizes the T2I base model to generate images as the first frame. In the qualitative experiments, for aesthetic purposes and to remove watermarks, an external model is used to generate a beautiful first frame.

\subsection{Qualitative Evaluation}
We choose AnimateDiff \cite{guo2023animatediff}, DynamiCrafter \cite{xing2023dynamicrafter}, and two commercialized products, Pika \cite{pikalab2023} and Gen2 \cite{gentwo2023}, for a composite qualitative analysis. It is worth noting that in the quantitative experiments, the first frame of FancyVideo is generated by SDXL to achieve a more aesthetically pleasing result and to minimize the appearance of watermark (although subsequent frames may still exhibit it).

As shown in Fig. \ref{Qualitative_eval}, our approach exhibits superior performance, outperforming previous methods regarding temporal consistency and motion richness.
In contrast, AnimateDiff, DynamiCrafter, and Gen2 generate videos with less motion.
Pika struggles to produce object-consistent and high-quality video frames.
Remarkably, our method can accurately understand the motion instructions in the text prompt (e.g., "A teddy bear walking ... beautiful sunset." and "A teddy bear running ... City." case).

\subsection{Quantitative Evaluation}
\label{Quantitative_Evaluation}

For a comprehensive comparison with the SOTA methods, we adopt three popular benchmarks (e.g., EvalCrafter \cite{liu2023evalcrafter}, UCF-101 \cite{soomro2012dataset}, and MSR-VTT \cite{xu2016msr} ) and human evaluation to evaluate the quality of video generation. Among them, EvalCrafter is a relatively comprehensive benchmark for video generation currently. UCF-101 and MSR-VTT are benchmarks commonly used in previous methods \cite{girdhar2023emu,zhang2023show}. Meanwhile, human evaluation can compensate for the inaccuracies in existing text-conditioned video generation evaluation systems.

\begin{table*}[htbp]
\centering
\resizebox{0.87\textwidth}{!}{
\begin{tabular}{cl|cccccc|c}

\hline
\multirow{2}{*}{Dimensions} & \multirow{2}{*}{Metrics}         & \multirow{2}{*}{Pika} & \multirow{2}{*}{Gen2} & \multirow{2}{*}{Show-1} & \multirow{2}{*}{Lumiere} & \multirow{2}{*}{DynamiCrafter} & \multirow{2}{*}{AnimateDiff} & \multirow{2}{*}{FancyVideo}    \\
                            &                                  &                                                          &                                                         &                                                              &                         &                           &                           &                                \\ \hline
                            & VQAA($\uparrow$)                 & 59.09                                                    & 59.44                                                   & 23.19                                                        & 40.06                                            & 74.56                                                & 65.94                                              & 85.78                          \\
Video                       & VQAT($\uparrow$)                 & 64.96                                                    & 76.51                                                   & 44.24                                                        & 32.93                                            & 59.48                                                & 52.02                                              & 74.56                          \\
Quality                     & IS($\uparrow$)                   & 14.81                                                    & 14.53                                                   & 17.65                                                        & 17.64                                            & 18.37                                                & 16.54                                              & 17.38                          \\
                            & Comprehensive($\uparrow$)        & 138.86                                                   & 150.48                                                  & 85.08                                                        & 90.63                                            & \underline{152.41}                      & 134.50                                             & \textbf{177.72}                \\ \hline
                            & CLIP-Score($\uparrow$)           & 20.46                                                    & 20.53                                                   & 20.66                                                        & 20.36                                            & 20.80                                                & 19.70                                              & 20.85                          \\
                            & BLIP-BLEU($\uparrow$)            & 21.14                                                    & 22.24                                                   & 23.24                                                        & 22.54                                            & 20.93                                                & 20.67                                              & 21.33                          \\
                            & SD-Score($\uparrow$)             & 68.57                                                    & 68.58                                                   & 68.42                                                        & 67.93                                            & 67.87                                                & 66.13                                              & 68.14                          \\
Text-Video                  & Detection-Score($\uparrow$)      & 58.99                                                    & 64.05                                                   & 58.63                                                        & 50.01                                            & 64.04                                                & 51.19                                              & 66.66                          \\
Alignment                   & Color-Score($\uparrow$)          & 34.35                                                    & 37.56                                                   & 48.55                                                        & 38.72                                            & 45.65                                                & 42.39                                              & 51.09                          \\
                            & Count-Score($\uparrow$)          & 51.46                                                    & 53.31                                                   & 44.31                                                        & 44.18                                            & 53.53                                                & 22.40                                              & 59.19                          \\
                            & OCR Score($\downarrow$)          & 84.31                                                    & 75.00                                                   & 58.97                                                        & 71.32                                            & 60.29                                                & 45.21                                              & 64.85                          \\
                            & Celebrity ID Score($\downarrow$) & 45.31                                                    & 41.25                                                   & 37.93                                                        & 44.56                                            & 26.35                                                & 42.26                                              & 25.76                          \\
                            & Comprehensive($\uparrow$)        & 325.35                                                   & 350.02                                                  & 366.91                                                       & 327.86                                           & \underline{386.18}                      & 335.01                                             & \textbf{396.65}                \\ \hline
                            & Action Score($\uparrow$)         & 71.81                                                    & 62.53                                                   & 81.56                                                        & 72.12                                            & 72.22                                                & 61.94                                              & 72.99                          \\
Motion                      & Motion AC-Score($\rightarrow$)   & 44                                                       & 44                                                      & 50                                                           & 42                                               & 46                                                   & 32                                                 & 52                             \\
Quality                     & Flow-Score($\rightarrow$)        & 0.50                                                     & 0.70                                                    & 2.07                                                         & 6.99                                             & 0.96                                                 & 2.403                                              & 1.7413                         \\
                            & Comprehensive($\uparrow$)        & 71.81                                                    & 62.53                                                   & \textbf{81.56}                                               & 72.12                                            & 72.22                                                & 61.94                                              & \underline{72.99} \\ \hline
                            & CLIP-Temp($\uparrow$)            & 99.97                                                    & 99.94                                                   & 99.77                                                        & 99.74                                            & 99.75                                                & 99.85                                              & 99.84                          \\
Temporal                    & Warping Error($\downarrow$)      & 0.0006                                                   & 0.0008                                                  & 0.0067                                                       & 0.0162                                           & 0.0054                                               & 0.0177                                             & 0.0051                         \\
Consistency                 & Face Consistency($\uparrow$)     & 99.62                                                    & 99.06                                                   & 99.32                                                        & 98.94                                            & 99.34                                                & 99.63                                              & 99.31                          \\
                            & Comprehensive($\uparrow$)        & \textbf{199.59}                                          & 199.00                                                  & 199.09                                                       & 198.68                                           & 199.09                                               & \underline{199.48}                    & 199.15                         \\ \hline
\end{tabular}}
  \caption{Quantitative evaluation on the EvalCrafter. The best and second performing metrics are highlighted in \textbf{bold} and \underline{underline}. Comprehensive denotes the composite metrics for these dimensions.}
\label{EvalCrafter-table}
\end{table*}

\paragraph{EvalCrafter Benchmark.}
\label{EvalCrafter Benchmark}
EvalCrafter \cite{liu2023evalcrafter} quantitatively evaluates the quality of text-to-video generation from four aspects (including Video Quality, Text-video Alignment, Motion Quality, and Temporal Consistency). Each dimension contains multiple subcategories of indicators shown in the Table. \ref{EvalCrafter-table}. 
As discussed in community \cite{evalcraftergithub2023}, the authors acknowledge that the original manner of calculating the comprehensive metric was inappropriate. For a more intuitive comparison, we introduce a comprehensive metric for every aspect by considering each sub-indicators numerical scale and positive-negative attributes. 

In detail, we compare the performance of the previous video generation SOTA methods (e.g., Pika \cite{pikalab2023}, Gen2 \cite{gentwo2023}, Show-1 \cite{zhang2023show}, Lumiere \cite{bar2024lumiere}, DynamiCrafter \cite{xing2023dynamicrafter}, and AnimateDiff \cite{guo2023animatediff}) and exhibit in Table. \ref{EvalCrafter-table}.  
Our method demonstrates outstanding performance beyond existing methods at the Video Quality and Text-video Alignment aspect.
Although Show-1 has the best Motion Quality (81.56), its Video Quality is poor (only 85.08). That indicates that it cannot generate high-quality videos with reasonable motion. 
However, our method has the second highest Motion Quality (72.99) and the best Video Quality (177.72), achieving the trade-off between quality and motion.
The above results indicate the superiority of FancyVideo and its ability to generate temporal-consistent and motion-accurate video.

\begin{table}
\renewcommand\arraystretch{0.94}
\setlength\tabcolsep{1.0pt}%
\centering
\small
{
{

\resizebox{0.44\textwidth}{!}{
  \begin{tabular}{cc|ccc|cc}
    \toprule[1.0pt]
    \multirow{2}{*}{Method} & \multirow{2}{*}{Data} & \multicolumn{3}{c|}{UCF-101} & \multicolumn{2}{c}{MSR-VTT} \\
    \cmidrule{3-7}
      &  & FVD($\downarrow$) & IS($\uparrow$) & FID($\downarrow$) & FVD($\downarrow$) & CLIPSIM ($\uparrow$) \\
    \midrule
    Emu Video & 34M  & 606.20  & \underline{42.70} & - & - & -    \\
    AnimateDiff  & 10M & 584.85  & 37.01 & 61.24 & 628.57 & 0.2881 \\
    DynamiCrafter & 10M & 404.50  & 41.97 & \textbf{32.35} & \textbf{219.31} & 0.2659  \\
    Show-1 & 10M & \underline{394.46}  & 35.42 & - & 538.00 & 0.3072   \\
    Lumiere & 10M & \textbf{332.49}  & 37.54 & - & 550.00 & 0.2939   \\
    \midrule
    FancyVideo & 10M & 412.64  & \textbf{43.66} & \underline{47.01} & \underline{333.52} & \textbf{0.3076}   \\
    \bottomrule[1.0pt]
  \end{tabular}
  }
}
}

\caption{Quantitative evaluation on the UCF-101 \protect\cite{soomro2012dataset} and MSR-VTT \protect\cite{xu2016msr} .The best and second performing metrics are highlighted in \textbf{bold} and \underline{underline} respectively.}

\label{UCF-MSR-table}
\end{table}
\paragraph{UCF-101 \& MSR-VTT.}

Following the prior work \cite{zhang2023show}, we evaluate the zero-shot generation performance on UCF-101 \cite{soomro2012dataset} and MSR-VTT \cite{xu2016msr} as shown in Table. \ref{UCF-MSR-table}. We use Fr\'echet Video Distance (FVD) \cite{unterthiner2019fvd}, Inception Score (IS) \cite{wu2021godiva}, Fr\'echet Inception Distance (FID) \cite{heusel2017gans}, and CLIP similarity (CLIPSIM) as evaluation metrics and compared some current SOTA methods. FancyVideo achieves competitive results, particularly excelling in IS and CLIPSIM with scores of 43.66 and 0.3076, respectively. Besides, previous studies \cite{ho2022imagen,girdhar2023emu,wu2023freeinit} have pointed out that these metrics do not accurately reflect human perception and are affected by the gap between the distribution of training and test data and the image's low-level detail.

\paragraph{Human Evaluation.}
\begin{figure}[!htp]
  \centering
  \includegraphics[width=0.96\linewidth]{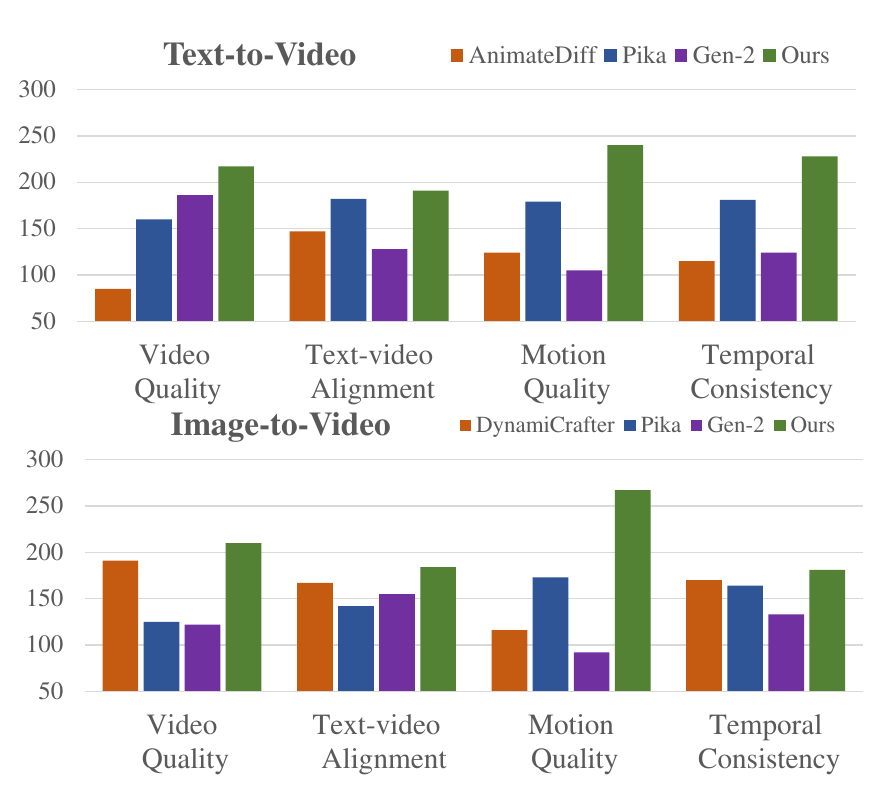}
    
  \caption{Human Evaluation Comparison. FancyVideo stands out significantly compared to other text-to-video and image-to-video generators in terms of Motion Quality and Temporal Consistency.}
  \label{human_eval}
\end{figure}
Inspired by EvalCrafter \cite{liu2023evalcrafter}, we introduce a multi-candidate ranking protocol with four aspects: video quality, text-video alignment, motion quality, and temporal consistency. In this protocol, participants rank the results of multiple candidate models for each aspect. Each candidate model receives a score based on its ranking. For instance, if there are $N$ candidate models ranked by video quality, the first model gets $N-1$ points, the second gets $N-2$ points, and so on, with the last model receiving 0 points. Adhering to this protocol, we selected 108 samples from the EvalCrafter validation set and gathered judgments from 100 individuals.
As depicted in Fig. \ref{human_eval}, our method significantly outperforms text-to-video conversion methods, including AnimateDiff \cite{guo2023animatediff}, Pika \cite{pikalab2023}, and Gen2 \cite{gentwo2023}, across all four aspects. FancyVideo demonstrates exceptional motion quality while preserving superior text-video consistency.
Additionally, we conducted a similar comparison of four image-to-video methods, including DynamiCrafter \cite{xing2023dynamicrafter}, Pika, and Gen2, as shown in Fig. \ref{human_eval}.

\subsection{Ablation Studies}

In this section, we conduct extensive experiments and exhibit detailed visual comparisons on the EvalCrafter benchmark \cite{liu2023evalcrafter} to thoroughly explore the effect of critical designs in CTGM. The ablation study includes two key modules (TII and TAR), each enhancing video quality As shown in Table \ref{Ablation-table}, TAR significantly improves both metrics, highlighting the importance of temporal attention refinement. Adding TII further enhances performance by refining latent features and enabling frame-level text control.

\begin{table}[]

\resizebox{0.48\textwidth}{!}{
\begin{tabular}{cc|cccc}
\hline
\multirow{2}{*}{TAR}      & \multirow{2}{*}{TII}      & Video                       & Text-Video                    & Motion                      & Temporal                        \\
                          &                           & \multicolumn{1}{l}{Quality} & \multicolumn{1}{l}{Alignment} & \multicolumn{1}{l}{Quality} & \multicolumn{1}{l}{Consistency} \\ \hline
                          &                           & 163.15                      & 361.92                        & 66.99                       & 198.83                          \\
\checkmark &                           & 172.44                      & 379.40                        & 71.24                       & 199.08                          \\
                          & \checkmark & 173.82                      & 380.24                        & 71.84                       & 199.04                          \\
\checkmark & \checkmark & 177.72                      & 396.65                        & 72.99                       & 199.15                          \\ \hline
\end{tabular}
}
\caption{Ablation studies on TAR and TII, where TFB is used by default in ablation experiments.}

\label{Ablation-table}
\end{table}

\section{Conclusion}
\label{sec:conclusion}

In this work, we present a novel video-generation method named FancyVideo, which optimizes common text control mechanisms (e.g., spatial cross-attention) from the cross-frame textual guidance. It improves cross-attention with a well-designed Cross-frame Textual Guidance Module (CTGM), implementing the temporal-specific textual condition guidance for video generation. A comprehensive qualitative and quantitative analysis shows it can produce more dynamic and consistent videos. This characteristic becomes more noticeable as the number of frames increases. Our method achieves state-of-the-art results on the EvalCrafter benchmark and human evaluations.


\twocolumn[{
\vspace*{1em}
\begin{center}
    {\LARGE \textbf{Supplementary Material for}}\\[1ex]
    {\LARGE FancyVideo: Towards Dynamic and Consistent Video Generation via Cross-frame Textual Guidance}\\[2ex]
\end{center}
\vspace{1em}
}]

\setcounter{section}{0}
\section{Appendix Section}
\label{sec:appendix_section}
\paragraph{Overview.}
In this supplemental material, we provide the following items:
\begin{itemize}
\item (Sec. 1) Temporal Feature Booster.
\item (Sec. 2) Experimental setup.
\item (Sec. 3) More details on evaluation metrics employed in UCF101, MSR-VTT, human evaluation and the EvalCrafter \cite{liu2023evalcrafter}.
\item (Sec. 4) More applications about Personalized Video Generation, high-resolution and multi-scale Video Generation, Video Predication and Video Backtracking, among others.
\item (Sec. 5) More results on video generation, including videos with different frame rates.
\item (Sec. 6) Prompts set for human evaluation.

\end{itemize}

\subsection{Temporal Feature Booster}
To further boost the temporal consistency of the feature, we process the $\mathcal{Z}_{ref}$ through the Temporal Feature Booster (TFB). This allows us to establish closer temporal connections. Specifically, TFB includes a simple yet effective temporal self-attention layer to refine the noisy latent feature along the time dimension, represented as:
\begin{equation}
\mathcal{Z'}_{ref} = \mathrm{TFB}(\mathcal{Z}_{ref}) = \mathrm{SelfAttn_t}(\mathcal{Z}_{ref}) + \mathcal{Z}_{ref}
\end{equation}

The ablation study of the Temporal Feature Booster (TFB) is presented in Table \ref{Ablation-tfb}.

\begin{table}[htbp]
\resizebox{0.48\textwidth}{!}{
\begin{tabular}{c|ccccc}
\hline
\multirow{2}{*}{TFB}      & Video                       & Text-Video                    & Motion                      & Temporal                        \\
                          & \multicolumn{1}{l}{Quality} & \multicolumn{1}{l}{Alignment} & \multicolumn{1}{l}{Quality} & \multicolumn{1}{l}{Consistency} \\ \hline
                          & 175.28                      & 391.21                        & 72.44                       & 199.05                          \\
\checkmark & 177.72                      & 396.65                        & 72.99                       & 199.15                          \\ \hline
\end{tabular}
}
\caption{Ablation studies on the TFB, where both TII and TAR are used by default in experimental configurations.}
\label{Ablation-tfb}
\end{table}

\subsection{Experimental Setup}
\label{Experimental_Setup}
\paragraph{Datasets.} We utilize WebVid-10M~\cite{bain2021frozen} as the training data. The WebVid-10M dataset contains 10.7 million video-caption pairs, with most videos having a resolution of 336 $\times$ 596. Since every clip in the WebVid-10M has a watermark, the generated video inevitably appears watermarked. 
\paragraph{Implementation details.}
FancyVideo is trained on the WebVid-10M dataset. The video clips are initially sampled with a stride of 4, followed by resizing and center-cropping to a resolution of 256 × 256. We utilize Stable-Diffusion v1.5 \cite{rombach2022high} as the text-to-image (T2I) base model and train exclusively with the temporal attention block and our CTGM block. For the 16-frame training, we use a batch size of 512 and train for 12,000 iterations. For training with 32 frames, 48 frames, and 64 frames, we use batch sizes of 256, 256, and 128, respectively. 
The training process comprises 24,000 iterations for 32 frames, 48,000 iterations for 48 frames, and 96,000 iterations for 64 frames on 64 A100 GPUs with 80G memory.
At inference, the sampling strategy for video generation is DDIM \cite{song2020denoising}
with 50 steps. Also, we utilize the classifier-free guidance \cite{ho2022classifier} with a 7.5 guidance scale. Similar to AnimateDiff \cite{guo2023animatediff}, we swap the base model with Realistic-Vision v5.1. The evaluations on EvalCrafter \cite{liu2023evalcrafter} and all the qualitative experiments are conducted at a resolution of 512. Regarding the evaluations on UCF-101 \cite{soomro2012dataset} and MSR-VTT \cite{xu2016msr}, following \cite{zhang2023show}, we conduct assessments on videos generated at a resolution of 256.

\subsection{More Details on Evaluation Metrics}
\label{metrices}
\subsubsection{Details of evaluation on UCF101}
To calculate the Fréchet Video Distance (FVD) \cite{unterthiner2019fvd} and Inception Score(IS) \cite{blattmann2023align,ren2024consisti2v,saito2020train}, we produce 2048 videos based on the class distribution of the UCF101 dataset. These videos are generated at a resolution of 256 pixels. Subsequently, we extract I3D embeddings from our videos. Next, we compute the FVD score by comparing the I3D embeddings of our videos with those of the UCF101 videos. For computing the Inception Score (IS), the same set of generated videos was utilized to extract C3D embeddings.

\subsubsection{Details of evaluation on MSR-VTT}
The MSR-VTT dataset is an open-domain video retrieval and captioning dataset with 10,000 videos, each having 20 captions. The standard splits include 6,513 training videos, 497 validation videos, and 2,990 test videos. For our experiments, we use the official test split and randomly select a text prompt for each video during evaluation. Using TorchMetrics, we compute our CLIPSIM \cite{wu2021godiva} metrics with the CLIP-VIT-B/32 model. We calculate the CLIP similarity for all frames in the generated videos and report the averaged results.

\subsubsection{Details of Human Evaluation}
We filter out 108 prompts from the 700 prompts in EvalCrafter for human evaluation. This subset of prompts contains more detailed descriptions. The prompt-108 we utilized is presented in \ref{prompt_list}.

\subsubsection{About EvalCrafter}
\label{EvalCrafter_metrics}
As mentioned in \ref{EvalCrafter Benchmark}, there are unreasonable aspects in how EvalCrafter calculates Video Quality, Text-Video Alignment, Motion Quality, and Temporal Consistency. Therefore, we propose our calculation scheme. Specifically, we remove some neutral sub-metrics and those with significant differences in scale, and based on whether each sub-metric is positive or negative, we obtain a comprehensive score. For Video Quality, it can be represented as follows:
\begin{equation}
Video\_Quality=VQAA+VQAT+IS
\end{equation}
As for Text-Video Alignment, it can be represented as:
\begin{figure*}[h]
  \centering
  \includegraphics[width=1\linewidth]{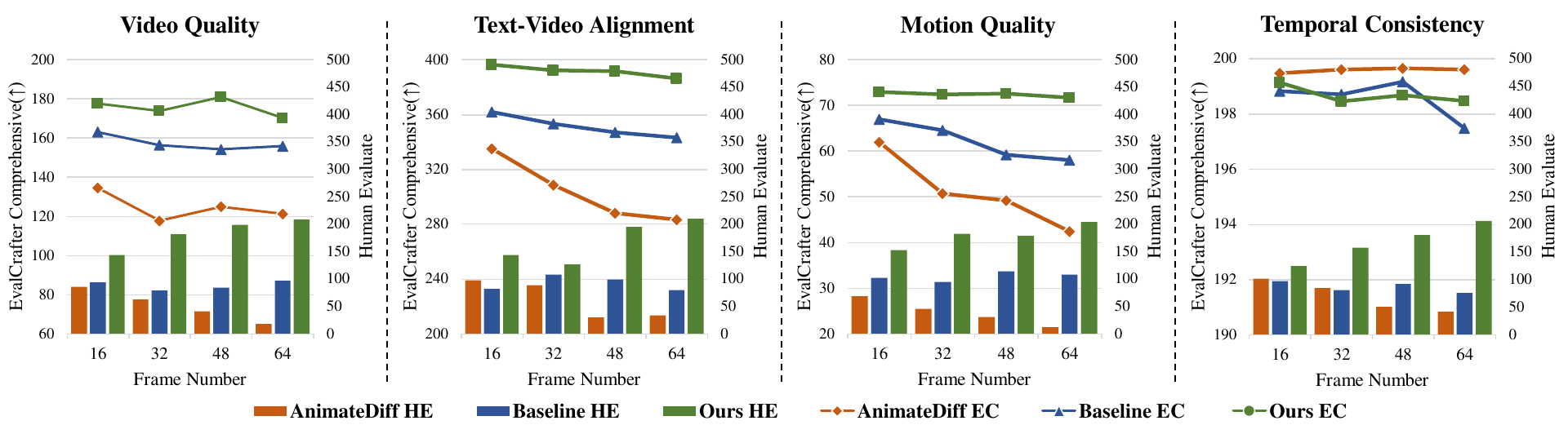}
  \caption{\textbf{Evaluation results of varying-frames video generation.} EC denotes the comprehensive metric on EvalCrafter benchmark and HE indicates Human Evaluation.}
  \label{long_video_eval}
\end{figure*}

\begin{align*}
Text\_Video\_Alignment &= CLIP\_Score +SD\_Score\\
                        &+BLIP\_BLEU  \\
&+ Count\_Score + Color\_Score\\
&+ Detection\_Score \\
&+ (100 - OCR\_Score) \\
&+ (100 - Celebrity\_ID\_Score)
\end{align*}

As for Motion Quality, we do not consider neutral metrics when calculating:
\begin{equation}
Motion\_Quality=Action\_Score
\end{equation}
As for Temporal Consistency, we neglect the warping error, which has a scale that differs significantly from other metrics:
\begin{equation}
Temporal\_Consistency=CLIP\_Temp+Face\_Consistency
\end{equation}
\begin{align*}
Temporal\_Consistency &= CLIP\_Temp\\
                        &+Face\_Consistency  \\
\end{align*}
\subsection{More Applications}
\label{more_app}
\begin{figure*}[h]
  \centering
  \includegraphics[width=1\linewidth]{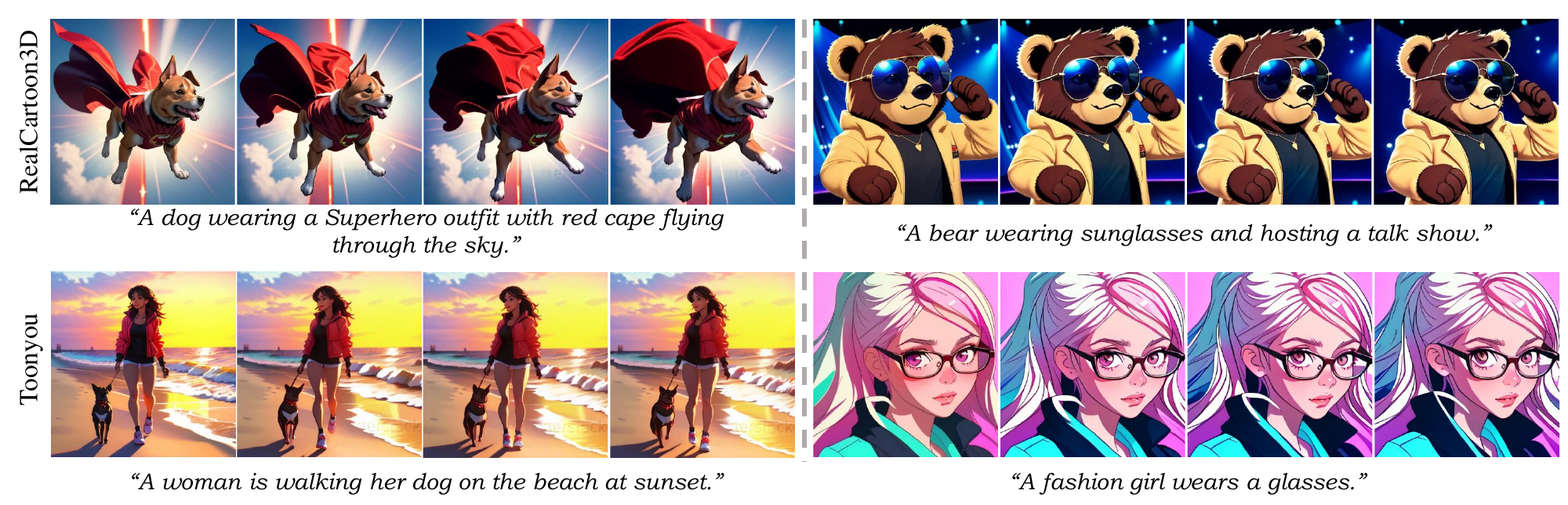}
    \vspace{-15pt}
  \caption{\textbf{Qualitative experiments.} We demonstrate the ability to generate personalized videos by using various T2I-based models. }
  \label{Personalize}
  \vspace{-10pt}
\end{figure*}

Given the flexibility of our method in swapping T2I base models, we conduct experiments similar to prior work \cite{guo2023animatediff}, with different base models. Fig. \ref{Personalize} displays the results of our method using models downloaded from the Civitai\cite{civitai2024} community. 

\begin{figure*}[htbp]
  \centering
  \includegraphics[width=1\linewidth]{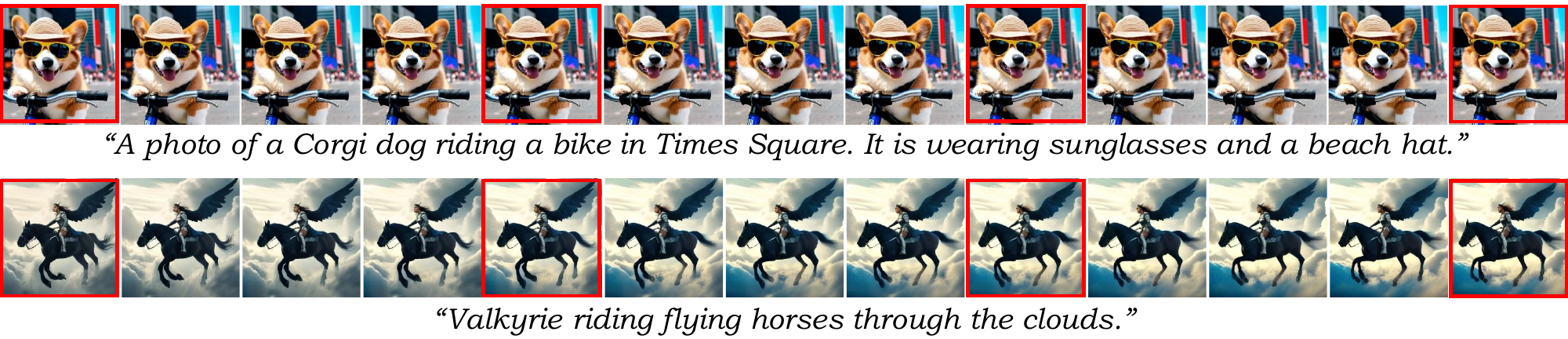}
    \vspace{-15pt}
  \caption{\textbf{Qualitative experiments.} Under the FancyVideo framework, we train the \textbf{Video Interpolation} model. The red border indicates the first four frames of the original video, and we inserted three frames between every two original frames.}
  \label{Video_Interpolation}
  \vspace{-5pt}
\end{figure*}

\begin{figure*}[h]
  \centering
  \includegraphics[width=1\linewidth]{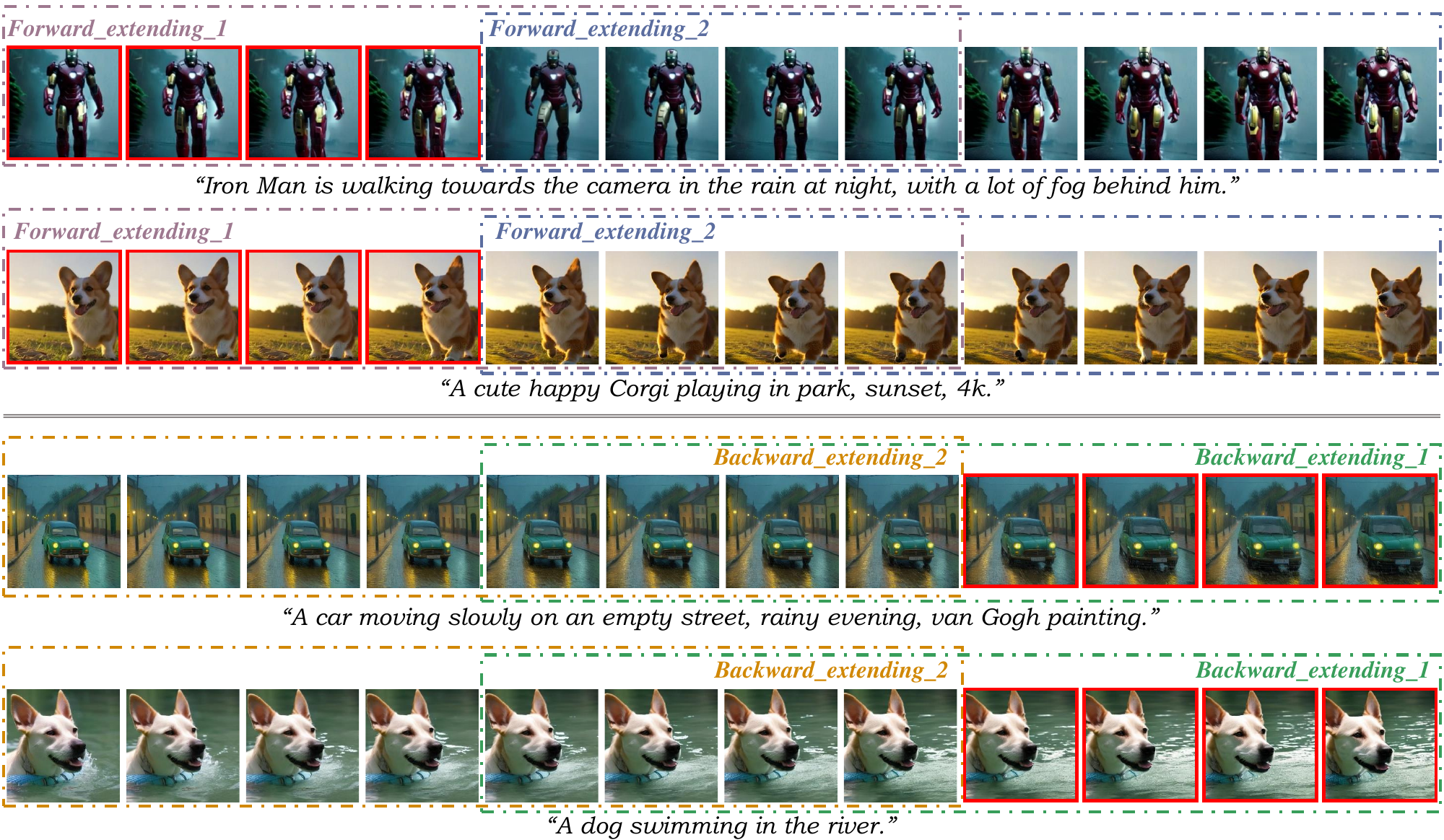}
    \vspace{-15pt}
  \caption{\textbf{Qualitative experiments.} Under the FancyVideo framework, we train the \textbf{Video Extending} models, which includes extending videos forward and extending videos backward. In the forward expansion model, the input consists of 4 frames in red border, and the output is the subsequent 4 frames. With two iterations, we extend a 4-frame video to 12 frames. \textbf{In the backward expansion model}, the input consists of 4 frames, and the output is the subsequent 4 frames. With two iterations, we extend a 4-frame video to 12 frames. Similarly, \textbf{in the backward expansion model}, the input consists of 4 frames in red border, and the output is the preceding 4 frames. We also perform two iterations.}
  \label{Video_extending_forward_backward}
  \vspace{-15pt}
\end{figure*}

Due to the specificity of our input, we investigate the potential of our model to perform additional functionalities by modifying the mask indicator and image indicator. This encompasses tasks such as Video Prediction and Video Extending, as depicted in Fig. \ref{Video_Interpolation} and Fig. \ref{Video_extending_forward_backward}.

Due to the flexibility of our framework, we can easily combine our method with text-to-image super-resolution modules \cite{cheng2024resadapter,xu2025dropoutgs,wang2025learning,wang2022spnet}. We conduct experiments to achieve multi-scale and high-resolution video generations, as depicted in Fig. \ref{Scale_and_solution}.

\subsection{More Generation Results}
\label{More_Generation_Results}
We further showcase additional results of the FancyVideo generation, including videos with 16 frames, 32 frames, 48 frames, and 64 frames. As shown in Fig. \ref{More_Results}, our method effectively maintains consistency while also addressing motion dynamics.

\begin{figure*}[htbp]
  \centering
  \includegraphics[width=1\linewidth]{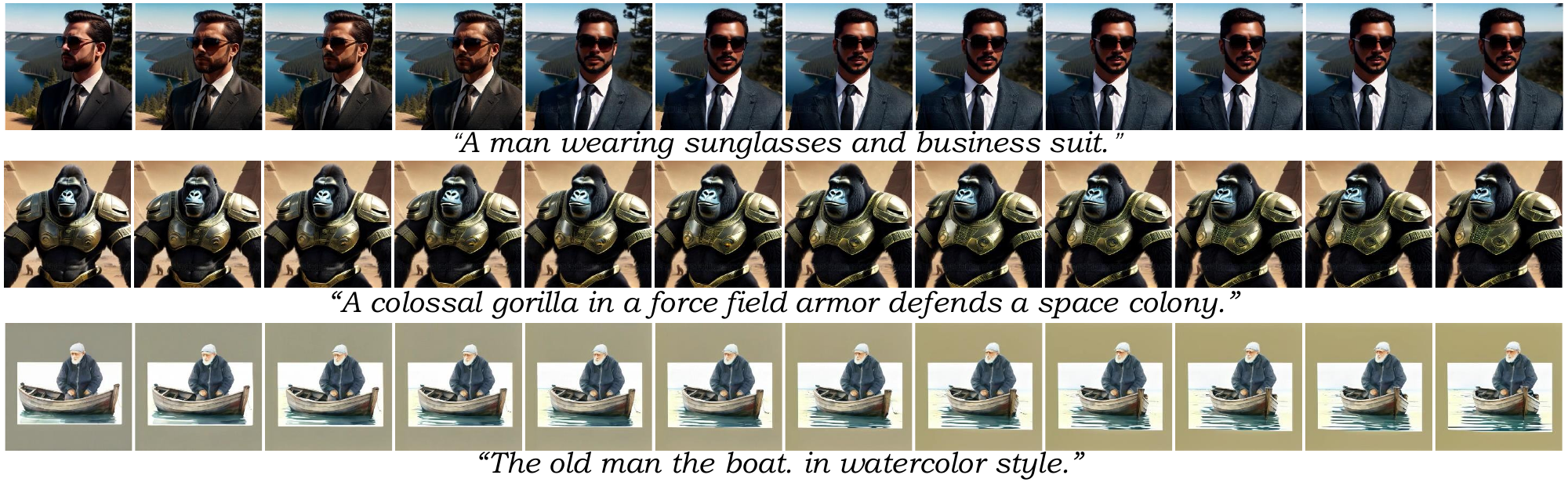}
  \caption{\textbf{Qualitative analysis} of our approach to long video generation (64 frames). More results are shown in \ref{More_Generation_Results}.}
  \label{long_video_vis_eval}
\end{figure*}
\begin{figure*}[htbp]
  \centering
  \includegraphics[width=0.95\linewidth]{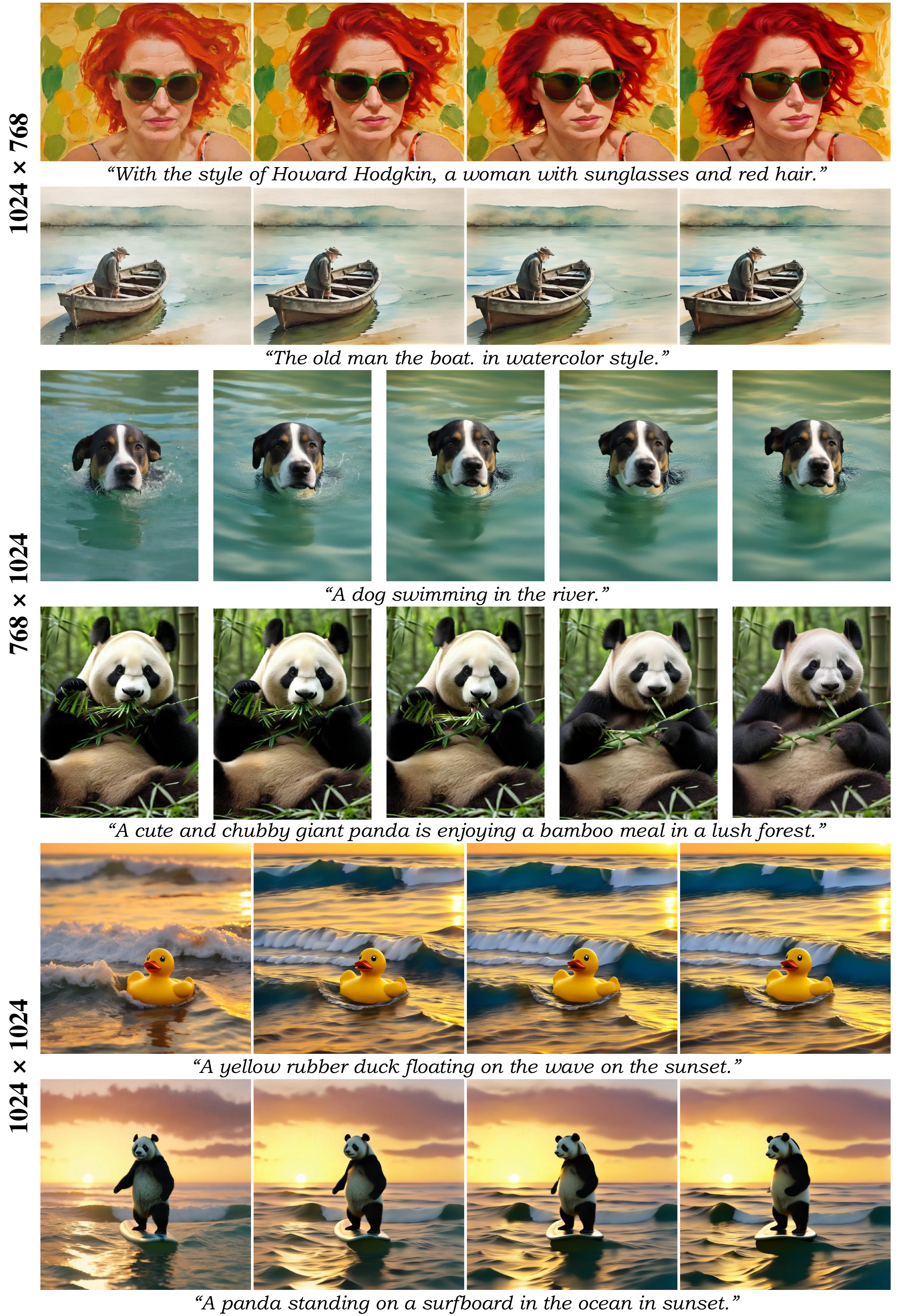}
    \vspace{-10pt}
  \caption{\textbf{Qualitative experiments.} By switching different base models, we demonstrate experimental results at multiple pixel scales. }
  \label{Scale_and_solution}
  \vspace{-5pt}
\end{figure*}

\begin{figure*}[htbp]
  \centering
  \includegraphics[width=1\linewidth]{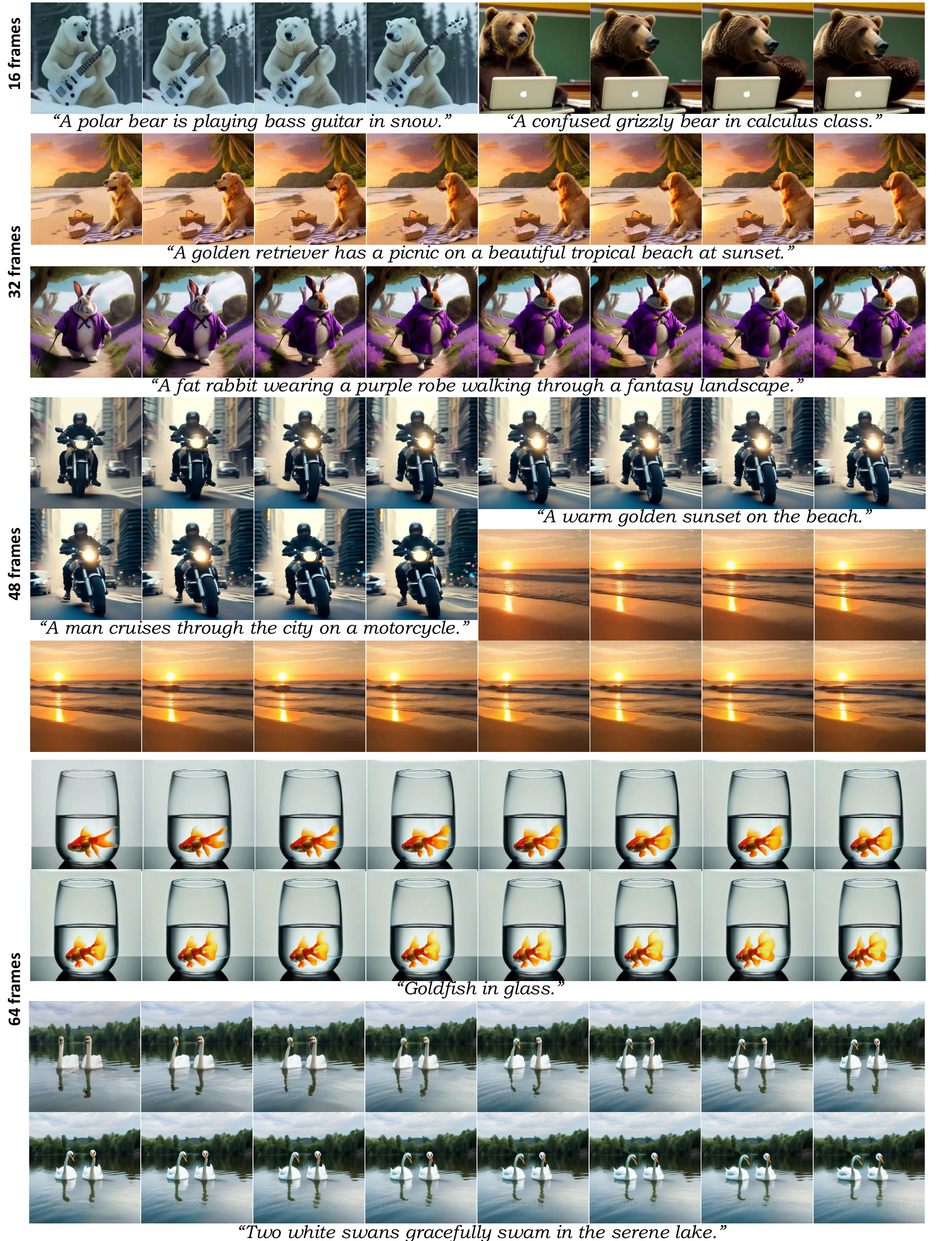}
    \vspace{-15pt}
  \caption{\textbf{Qualitative experiments.} We demonstrate the experimental results of FancyVideo with 16, 32, 48, and 64 frames.}
  \label{More_Results}
  \vspace{-15pt}
\end{figure*}

\subsection{A list of prompts for human evaluation}
\label{prompt_list}
\begin{itemize}
\item goldfish in glass
\item A peaceful cow grazing in a green field under the clear blue sky
\item A fluffy grey and white cat is lazily stretched out on a sunny window sill, enjoying a nap after a long day of lounging.
\item a horse
\item Two elephants are playing on the beach and enjoying a delicious beef stroganoff meal.
\item A slithering snake moves through the lush green grass
\item A cute and chubby giant panda is enjoying a bamboo meal in a lush forest. The panda is relaxed and content as it eats, and occasionally stops to scratch its ear with its paw.
\item Pikachu snowboarding
\item a dog wearing vr goggles on a boat
\item In an African savanna, a majestic lion is prancing behind a small timid rabbit. The rabbit tried to run away, but the lion catches up easily. 
\item A photo of a Corgi dog riding a bike in Times Square. It is wearing sunglasses and a beach hat.
\item In the lush forest, a tiger is wandering around with a vigilant gaze while the birds chirp and monkeys play.
\item A family of four fluffy, blue penguins waddled along the icy shore.
\item Two white swans gracefully swam in the serene lake
\item A bear rummages through a dumpster, searching for food scraps.
\item light wind, feathers moving, she moves her gaze, 4k 
\item fashion portrait shoot of a girl in colorful glasses, a breeze moves her hair 
\item Two birds flew around a person, in the style of Sci-Fi
\item flying superman, hand moves forward
\item Batman turns his head from right to left
\item Iron Man is walking towards the camera in the rain at night, with a lot of fog behind him. Science fiction movie, close-up
\item Bruce Lee shout like a lion ,wild fighter
\item A woman is walking her dog on the beach at sunset.
\item Valkyrie riding flying horses through the clouds
\item A surfer paddles out into the ocean, scanning the waves for the perfect ride.
\item A man cruises through the city on a motorcycle, feeling the adrenaline rush
\item A musician strums his guitar, serenading the moonlit night
\item Leaves falling in autumn forest
\item Thunderstorm at night
\item A snow avalanche crashed down a mountain peak, causing destruction and mayhem
\item A thick fog covers a street, making it nearly impossible to see. Cars headlights pierce through the mist as they slowly make their way down the road.
\item The flowing water sparkled under the golden sunrise in a peaceful mountain river.
\item A warm golden sunset on the beach, with waves gently lapping the shore.
\item Mount Fuji
\item A beautiful leather handbag caught my eye in the store window. It had a classic shape and was a rich cognac color. The material was soft and supple. The golden text label on the front read 'Michael Kors'.
\item balloons flying in the air
\item a motorcycle race through the city streets at night
\item A silver metal train with blue and red stripes, speeding through a mountainous landscape.
\item hot ramen
\item Juicy and sweet mangoes lying in a woven basket
\item A delicious hamburger with juicy beef patty, crispy lettuce and melted cheese.
\item In Marvel movie style, supercute siamese cat as sushi chef
\item With the style of Egyptian tomp hieroglyphics, A colossal gorilla in a force field armor defends a space colony.
\item a moose with the style of Hokusai
\item a cartoon pig playing his guitar, Andrew Warhol style
\item A cat watching the starry night by Vincent Van Gogh, Highly Detailed, 2K with the style of emoji
\item impressionist style, a yellow rubber duck floating on the wave on the sunset
\item A Egyptian tomp hieroglyphics painting ofA regal lion, decked out in a jeweled crown, surveys his kingdom.
\item Macro len style, A tiny mouse in a dainty dress holds a parasol to shield from the sun.
\item A young woman with blonde hair, blue eyes, and a prominent nose stands at a bus stop in a red coat, checking her phone. in the style of Anime, anime style
\item pikachu jedi, film realistic, red sword in renaissance style style
\item abstract cubism style, Freckles dot the girl's cheeks as she grins playfully
\item Howard Hodgkin style, A couple walks hand in hand along a beach, watching the sunset as they talk about their future together.
\item A horse sitting on an astronaut's shoulders. in Andrew Warhol style
\item With the style of Howard Hodgkin, a woman with sunglasses and red hair
\item The old man the boat. in watercolor style
\item One morning I chased an elephant in my pajamas, Disney movie style
\item A rainbow arched across the sky, adding a burst of color to the green meadow.  in Egyptian tomp hieroglyphics style
\item In Roy Lichtenstein style, In the video, a serene waterfall cascades down a rocky terrain. The water flows gently, creating a peaceful ambiance.
\item The night is dark and quiet. Only the dim light of streetlamps illuminates the deserted street. The camera slowly pans across the empty road. with the style of da Vinci
\item New York Skyline with 'Hello World' written with fireworks on the sky. in anime style
\item A car on the left of a bus., oil painting style
\item traditional Chinese painting style, a pickup truck at the beach at sunrise
\item a sword, Disney movie style
\item a statue with the style of van gogh
\item orange and white cat., slow motion
\item slow motion, A brown bird and a blue bear.
\item Two elephants are playing on the beach and enjoying a delicious beef stroganoff meal., camera rotate anticlockwise
\item camera pan from right to left, A trio of powerful grizzly bears fishes for salmon in the rushing waters of Alaska
\item a Triceratops charging down a hill, camera pan from left to right
\item hand-held camera, A real life photography of super mario, 8k Ultra HD.
\item drove viewpoint, a man wearing sunglasses and business suit
\item a girl with long curly blonde hair and sunglasses, camera pan from left to right
\item close-up shot, high detailed, a girl with long curly blonde hair and sunglasses
\item an old man with a long grey beard and green eyes, camera rotate anticlockwise
\item camera pan from left to right, a smiling man
\item drove viewpoint, fireworks above the Parthenon
\item zoom in, A serene river flowing gently under a rustic wooden bridge.
\item large motion, a flag with a dinosaur on it
\item still camera, an F1 race car
\item hand-held camera, Three-quarters front view of a blue 1977 Corvette coming around a curve in a mountain road and looking over a green valley on a cloudy day.
\item drove viewpoint, wine bottles
\item a paranoid android freaking out and jumping into the air because it is surrounded by colorful Easter eggs, camera rotate anticlockwise
\item A traveler explores a scenic trail on the back of a sturdy mule, taking in the breathtaking views of the mountains.
\item A farmer drives a tractor through a vast field, tending to the crops with care and expertise.
\item A violinist moves her bow in a large motion sweep, creating a beautiful melody.
\item A swimmer dives into the water with a large motion splash, beginning a race.
\item A drummer hits the cymbals with a large motion crash, punctuating the music.
\item Slow motion raindrops fall gently from the sky, creating ripples in a puddle.
\item Slow motion leaves fall from a tree, swirling through the air.
\item Slow motion lightning illuminates the dark sky, followed by the rumble of thunder.
\item Slow motion bubbles rise to the surface of a glass of champagne.
\item Slow motion smoke curls up from a burning candle.
\item Slow motion confetti falls from the sky, celebrating a victory.
\item Slow motion steam rises from a hot cup of coffee.
\item Slow motion birds soar through the sky, their wings outstretched.
\item Three dogs playfully chase each other around a park.
\item Three horses gallop across a wide open field, tails and manes flying in the wind.
\item A blue boat sailing on the water with a red flag.
\item A person riding a green motorbike with an orange helmet.
\item A green cow grazing in a field with a yellow sun.
\item A yellow cat sleeping on a green bench.
\item Brad Pitt smirks charmingly, his blue eyes sparkling with mischief.
\item Angelina Jolie's full lips curve into a smile, her gaze intense and captivating.
\item Tom Cruise's intense stare conveys determination, his jaw set firmly.
\item Leonardo DiCaprio's eyes glimmer with passion, his handsome face displaying intensity.
\item Robert Downey Jr.'s smug grin conveys his character's confidence, his eyes full of wit.
\item Johnny Depp's face shows playfulness, his eyes twinkling with mischief.
\end{itemize}

\bibliographystyle{named}
\bibliography{ijcai25}

\end{document}